\documentclass[11pt]{article}

\usepackage[preprint]{acl}

\usepackage{times}
\usepackage{latexsym}

\usepackage[T1]{fontenc}

\usepackage[utf8]{inputenc}

\usepackage{microtype}

\usepackage{inconsolata}

\usepackage{graphicx}

\usepackage{amsmath,amssymb,amsfonts}
\usepackage{booktabs}
\usepackage{algorithm}
\usepackage{algorithmic}
\usepackage{multirow}
\usepackage{subcaption}
\usepackage[breakable]{tcolorbox}
\usepackage{enumitem}
\usepackage{placeins}

\providecommand{\linenumbers}{}
\providecommand{\nolinenumbers}{}

\newcommand{\method}{SAW}
\newcommand{\methodfull}{Stage-Aware Dynamic Weighting}
\newcommand{\cv}{\mathrm{CV}}

\title{SAW: Stage-Aware Dynamic Weighting for Multi-Objective \\ Reinforcement Learning in Large Language Models}

\author{
  Yuchen He$^{1,\ast}$, Baolong Bi$^{1,\ast}$, Shenghua Liu$^{1,\dagger}$, Huaming Liao$^{1,\dagger}$, Yuyao Ge$^{1}$ \\
  {\large\textbf{Bolin Wan$^{2}$, Siqian Tong$^{1}$, Juan Chen$^{1}$, Jiafeng Guo$^{1}$, Xueqi Cheng$^{1}$}} \\[4pt]
  $^{1}$Institute of Computing Technology, Chinese Academy of Sciences \\
  $^{2}$University of Electronic Science and Technology of China
}

\begin{document}
\maketitle

\renewcommand{\thefootnote}{\fnsymbol{footnote}}
\footnotetext[1]{Equal contribution.}
\footnotetext[2]{Corresponding authors.}
\renewcommand{\thefootnote}{\arabic{footnote}}

\begin{abstract}
Although multi-objective reinforcement learning (MORL) is central to aligning large language models with complex human preferences, the prevailing practice of static weighted summation overlooks a more fundamental phenomenon: \emph{reward learning is markedly asynchronous across objectives}. Well-learned dimensions quickly produce homogeneous, low-variance signals whose residual noise contaminates the aggregated reward (in GRPO) or occupies a fixed share of the advantage budget (in GDPO), interfering with the scarce yet high-value signals carried by under-learned dimensions. To address this asynchrony, we propose \textbf{Stage-Aware Dynamic Weighting} (\method{}), a lightweight, algorithm-agnostic dynamic weighting mechanism. \method{} utilizes the \textbf{coefficient of variation} (CV) as a scale-invariant proxy for real-time informativeness, reweighting each dimension's reward or advantage contribution by its relative informativeness within the batch. Unlike gradient-based methods that require multiple forward and backward passes, \method{} relies solely on batch-level statistics, introducing nearly negligible computational overhead. Experiments on tool-calling and text summarization tasks demonstrate that \method{} consistently improves both training efficiency and final performance under both GRPO and GDPO frameworks, confirming it as a general-purpose plug-in for multi-reward LLM alignment. Our code is available at \url{https://github.com/Zhaolutuan/SAW}.
\end{abstract}

\section{Introduction}
\label{sec:intro}

\nolinenumbers
\begin{figure*}[t]
    \centering
    \includegraphics[width=\textwidth]{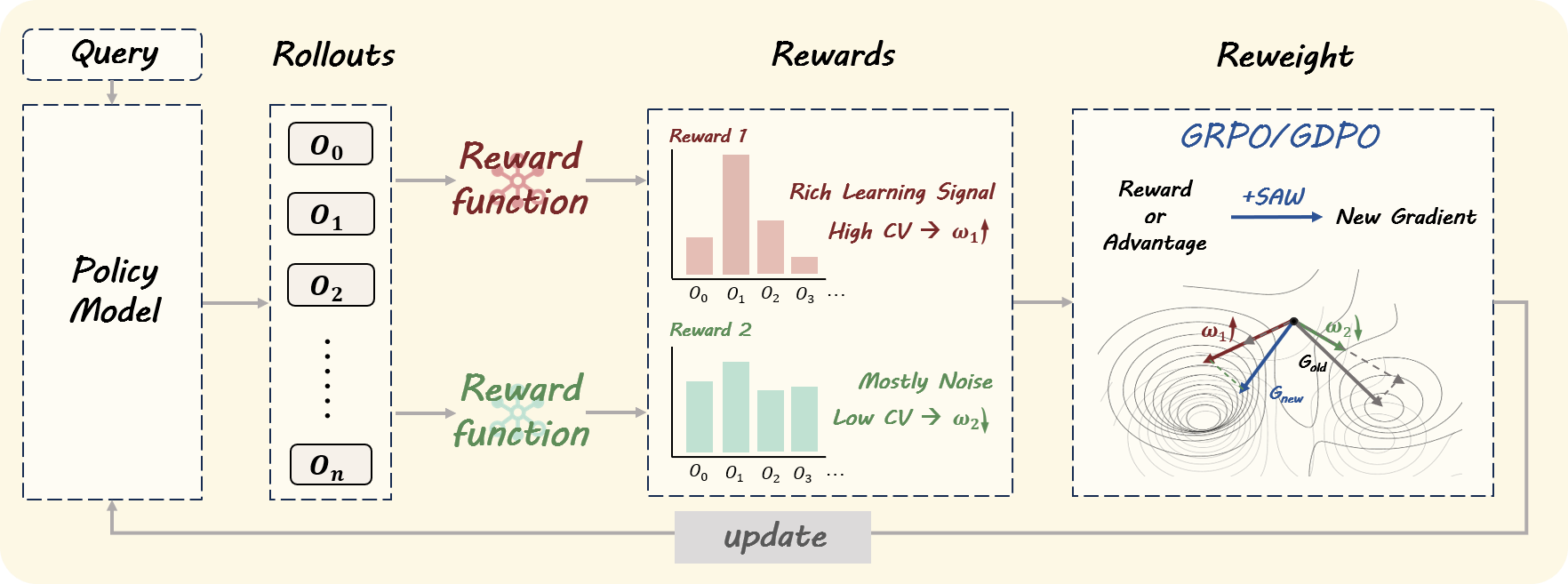}
    \caption{Overview of \method{}. Given a query, the policy model produces a group of rollouts that are scored by multiple reward functions. Within each batch, the resulting per-dimension reward distributions exhibit different degrees of dispersion: the dimension currently carrying a more informative signal (high CV) is upweighted ($\omega_1 \uparrow$), while the one whose batch signal has become more uniform and noise-dominated (low CV) is downweighted ($\omega_2 \downarrow$). \emph{The contour plot in the Reweight panel is a schematic 2D projection of the high-dimensional optimization landscape}: under static weighting, the noise-contaminated gradient $G_{\text{old}}$ steers the policy toward a sub-optimal basin, whereas SAW's CV-based reweighting yields a corrected gradient $G_{\text{new}}$ aligned with the true descent direction.}
    \label{fig:overview}
\end{figure*}
\linenumbers

\nolinenumbers
\begin{figure*}[t]
    \centering
    \includegraphics[width=\textwidth]{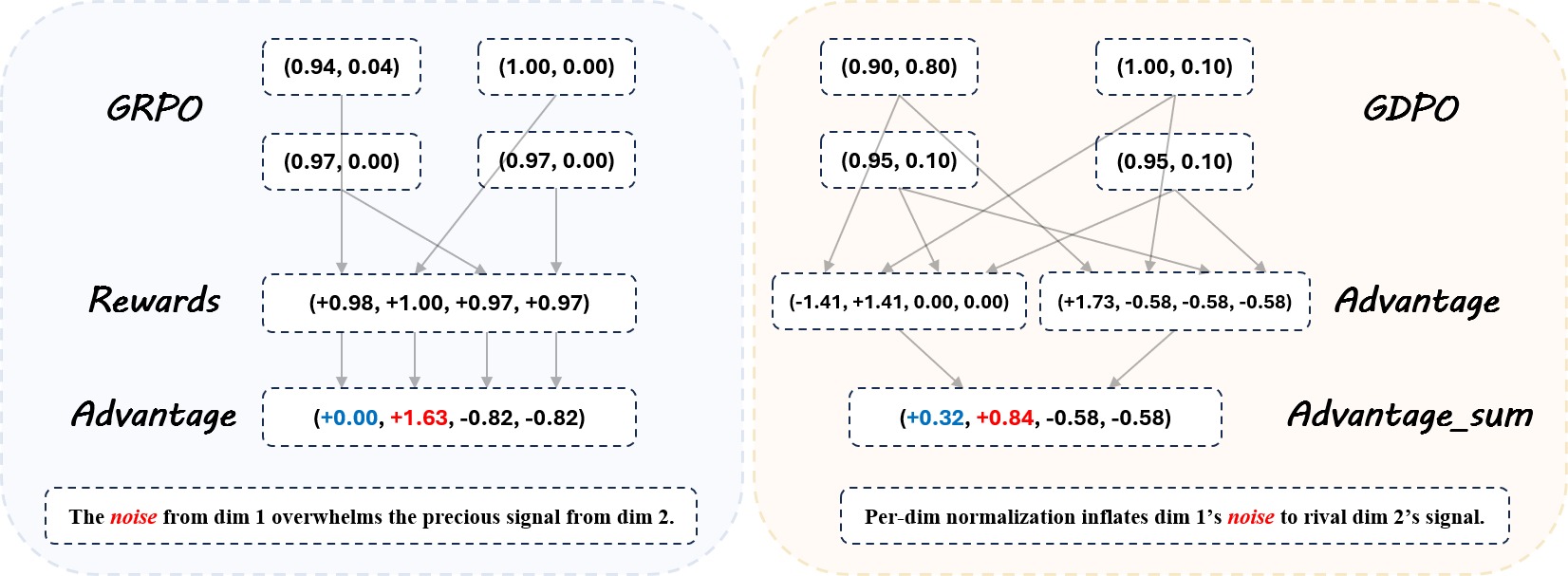}
    \caption{Two failure modes of static-weight aggregation under asynchronous reward learning, on minimal four-rollout examples in $[0,1]$. Both failures share the same root cause: static weights cannot tell collapsed noise from genuine signal. The detailed mechanism is analyzed in \S\ref{sec:noise_injection}.}
    \label{fig:noise_injection}
\end{figure*}
\linenumbers

Aligning large language models (LLMs) with human preferences \citep{christiano2017preferences, ouyang2022training, bai2022hh, rafailov2023dpo, lee2024rlaif, yuan2024selfrewarding} increasingly requires optimizing multiple objectives simultaneously.
Modern RL-based alignment pipelines incorporate diverse reward signals spanning correctness and reasoning ability \citep{deepseekr1}, efficiency \citep{kimi2025, liu2025adaptive, aggarwal2025l1}, and safety \citep{dai2023safe, bai2022cai, glaese2022sparrow}.
The prevailing practice aggregates these signals via static weighted summation, holding each reward dimension's weight fixed throughout training.
Yet this seemingly straightforward approach masks a deeper dynamic imbalance that fundamentally limits training effectiveness in multi-reward RL.

Recent work has begun to expose the limitations of naive reward aggregation schemes.
\citet{liu2025gdpo} demonstrate that applying group-wise normalization to the summed reward causes distinct reward combinations to collapse into identical advantage values, termed \emph{reward collapse}, and propose GDPO to resolve this problem by normalizing each reward dimension independently.

We identify a more fundamental problem that persists even under GDPO's improved normalization: \textbf{reward learning is asynchronous across objectives}.
Different reward objectives are mastered at different rates as training proceeds. This asynchrony manifests directly in the reward statistics: faster-learned dimensions quickly produce homogeneous, low-variance signals whose residual fluctuations behave as noise, while still-learning dimensions remain dispersed and continue to carry rich learning signal.
Under static aggregation, this noise from saturated dimensions is mixed into the signal that drives policy updates: in GRPO, the residual noise of saturated dimensions is summed into $r_{\text{sum}}$ before group normalization, contaminating the aggregated advantage and biasing the direction of every gradient update; in GDPO, where per-dimension advantages are normalized to the same scale, each saturated dimension still occupies a $1/n$ share of the advantage budget regardless of its information content, similarly biasing the gradient direction.
In both cases, the effective signal-to-noise ratio of the gradient direction degrades over training: scarce yet high-value signals from under-learned dimensions are interfered with by, or forced to compete on equal footing with, residual noise from already-saturated ones.

A natural response is to dynamically adjust the reward weights based on the training progress.
However, existing dynamic weighting methods from multi-task learning (such as gradient-based approaches \citep{sener2018mgda, pcgrad, cagrad, nashmtl, chen2018gradnorm} that derive weights from per-objective gradient interactions) require multiple additional forward and backward passes per training step.
In the LLM setting, where a single forward-backward pass is already expensive, such overhead is prohibitive for online RL training at scale.

To address this problem, we propose \textbf{\methodfull{}} (\method{}; Figure~\ref{fig:overview}), a lightweight, algorithm-agnostic dynamic weighting mechanism.
\method{} utilizes the \textbf{coefficient of variation} (CV), defined as the ratio of the standard deviation to the mean, as a scale-invariant proxy for the real-time informativeness of each reward dimension.
A higher CV indicates that, within the current batch, the dimension's rewards remain dispersed and carry an informative learning signal; a lower CV indicates that its rewards have become more uniform and the information it provides has diminished.
By making weights proportional to each dimension's CV share, \method{} reallocates each dimension's weight at the reward level (GRPO) or the advantage level (GDPO) according to its relative informativeness. This amplifies the dimension carrying a scarcer, more informative signal while proportionally downweighting the one whose batch signal has become more uniform, using only batch-level mean and standard deviation, with no gradient computation or Pareto front estimation.

Overall, our contributions are as follows:

\begin{itemize}[leftmargin=*, itemsep=2pt, topsep=2pt]
    \item We identify and formalize the asynchrony in how distinct reward objectives are learned in multi-reward RL: under static weighting, the residual fluctuations of saturated dimensions corrupt the aggregated signal that drives the gradient update, while scarce, high-value signals from under-learned dimensions are systematically diluted and progressively obscured during training.
    \item We propose Stage-Aware Dynamic Weighting (\method{}), a statistics-driven dynamic weighting method that requires only batch-level mean and standard deviation, with no modification to the underlying reward functions or optimizer architecture, and serves as a modular plug-in applicable at the reward level with GRPO or the advantage level with GDPO.
    \item We evaluate \method{} on tool calling (ToolRL, with two model scales: 1.5B and 3B) and Reddit summarization (1.5B) under both GRPO and GDPO, demonstrating consistent improvements across tasks with different reward structures and markedly different reward dynamics.
\end{itemize}

\section{Preliminaries}
\label{sec:prelim}

\subsection{Multi-Objective Reinforcement Learning}

In multi-reward RL for LLMs, the training signal comprises $n$ reward functions $r_1, r_2, \ldots, r_n$, each evaluating a distinct quality dimension of model outputs (e.g., correctness, formatting, safety).
Given a question $q_i$ and a sampled response $o_j$, the aggregated training reward is computed as a weighted sum across dimensions:
\begin{equation}
    r_{\text{sum}}^{(i,j)} = \sum_{k=1}^{n} \omega_k \, r_k^{(i,j)}
    \label{eq:grpo_sum}
\end{equation}
where $\omega_k$ denotes the weight assigned to the $k$-th reward dimension.
In standard practice, these weights are set statically.

A key statistic used extensively throughout this paper is the \textbf{coefficient of variation} (CV). For reward dimension $k$ within a single batch of $N$ sampled responses, the CV is defined as:
\begin{equation}
    \cv_k = \frac{\sigma_k}{\mu_k + \delta}
    \label{eq:cv}
\end{equation}
where $\sigma_k$ and $\mu_k$ are the batch-level standard deviation and mean of $r_k$, respectively, and $\delta$ is a small constant for numerical stability.
Unlike raw variance, CV normalizes by the mean, making it a \emph{scale-invariant} measure of relative dispersion that enables meaningful comparison across reward dimensions with different numerical scales.

\subsection{RL Algorithms}

\paragraph{GRPO.} Group Relative Policy Optimization \citep{deepseekmath} eliminates the need for a value model by computing advantages relative to a group of sampled responses.
For a given question $q_i$, the behavior policy $\pi_{\theta_{\text{old}}}$ generates a group of $G$ responses $\{o_j\}_{j=1}^{G}$. The group-relative advantage is computed by normalizing the aggregated reward (Eq.~\ref{eq:grpo_sum}) within the same sampled group:
\begin{equation}
    A_{\text{sum}}^{(i,j)} = \frac{r_{\text{sum}}^{(i,j)} - \text{mean}\{r_{\text{sum}}^{(i,\cdot)}\}}{\text{std}\{r_{\text{sum}}^{(i,\cdot)}\}}
    \label{eq:grpo_adv}
\end{equation}
The policy is then updated by maximizing the clipped surrogate objective:
\begin{equation}
\begin{split}
    \mathcal{J}_{\text{GRPO}}(\theta) = \mathbb{E} \bigg[ \frac{1}{G} \sum_{j=1}^{G} \frac{1}{|o_j|} \sum_{t=1}^{|o_j|} \min\big( s_{j,t}\, A_{\text{sum}}^{(i,j)},\\
    \text{clip}(s_{j,t}, 1{-}\epsilon, 1{+}\epsilon)\, A_{\text{sum}}^{(i,j)} \big) \bigg]
\end{split}
    \label{eq:grpo_obj}
\end{equation}
where $s_{j,t} = \frac{\pi_\theta(o_j^t \mid q, o_j^{<t})}{\pi_{\theta_{\text{old}}}(o_j^t \mid q, o_j^{<t})}$ is the standard per-token importance sampling ratio.

\paragraph{GDPO.} Recently, \citet{liu2025gdpo} identify that directly applying GRPO's group-wise normalization to the summed reward causes \emph{reward collapse} (distinct reward combinations collapse into identical advantage values), and propose GDPO, which normalizes each individual reward dimension independently before final aggregation:
\begin{equation}
    A_k^{(i,j)} = \frac{r_k^{(i,j)} - \text{mean}\{r_k^{(i,\cdot)}\}}{\text{std}\{r_k^{(i,\cdot)}\}}, \; k = 1, \ldots, n
    \label{eq:gdpo_per_reward}
\end{equation}
\begin{equation}
    A_{\text{sum}}^{(i,j)} = A_1^{(i,j)} + A_2^{(i,j)} + \cdots + A_n^{(i,j)}
    \label{eq:gdpo_sum}
\end{equation}
followed by an additional batch-wise normalization to stabilize the overall scale:
\begin{equation}
    \hat{A}_{\text{sum}}^{(i,j)} = \frac{A_{\text{sum}}^{(i,j)} - \text{mean}\{A_{\text{sum}}^{(\cdot)}\}}{\text{std}\{A_{\text{sum}}^{(\cdot)}\} + \delta}
    \label{eq:gdpo_batchnorm}
\end{equation}

\begin{table*}[t]
    \centering
    \small
    \begin{tabular}{l @{\hskip 3.0em} c c c c}
        \toprule
        \textbf{Method} & \textbf{Overall Acc $\uparrow$} & \textbf{Non-Live Overall Acc $\uparrow$} & \textbf{Live Overall Acc $\uparrow$} & \textbf{Multi-Turn Overall Acc $\uparrow$} \\
        \midrule
        \multicolumn{5}{l}{\textit{Qwen2.5-1.5B-Instruct}} \\
        \midrule
        Base              & 31.25\%          & 30.92\%          & 48.78\%          & 2.13\%  \\
        GRPO              & 32.55\%          & 32.81\%          & 50.26\%          & 2.25\%  \\
        GRPO+Gradient     & 33.41\%          & \textbf{33.98\%} & 51.22\%          & 2.50\%  \\
        GRPO+\method{}    & \textbf{33.53\%} & 33.10\%          & \textbf{52.19\%} & \textbf{2.63\%}  \\
        \midrule
        GDPO              & 33.82\%          & \textbf{34.29\%} & 51.96\%          & 2.50\%  \\
        GDPO+Gradient     & 33.93\%          & 33.48\%          & \textbf{52.78\%} & \textbf{2.75\%} \\
        GDPO+\method{}    & \textbf{34.02\%} & 33.98\%          & 52.63\%          & 2.63\%  \\
        \midrule
        \multicolumn{5}{l}{\textit{Qwen2.5-3B-Instruct}} \\
        \midrule
        Base              & 34.57\%          & 35.13\%          & 52.19\%          & 4.00\%  \\
        GRPO              & 36.27\%          & 35.94\%          & \textbf{55.29\%} & 4.63\%  \\
        GRPO+Gradient     & 35.60\%          & 35.75\%          & 54.56\%          & 3.38\%  \\
        GRPO+\method{}    & \textbf{36.35\%} & \textbf{36.06\%} & 55.22\%          & \textbf{4.88\%} \\
        \midrule
        GDPO              & 35.58\%          & 35.71\%          & 54.03\%          & 4.25\%  \\
        GDPO+Gradient     & 36.43\%          & \textbf{36.31\%} & 55.22\%          & \textbf{4.88\%} \\
        GDPO+\method{}    & \textbf{36.62\%} & 35.98\%          & \textbf{56.40\%} & 4.12\%  \\
        \bottomrule
    \end{tabular}
    \caption{ToolRL accuracy on BFCL-v4 (averaged over 4 evaluation runs). Bold indicates the best result within each model and algorithm family (GRPO-based or GDPO-based).}
    \label{tab:toolrl}
\end{table*}

\section{Motivation}
\label{sec:motivation}

To empirically ground the asynchronous reward learning phenomenon described in \S\ref{sec:intro}, this section first observes the phenomenon on a real training run (\S\ref{sec:async_curves}), then uses a concrete four-rollout example to show how the resulting noise is injected into GRPO and GDPO policy updates respectively (\S\ref{sec:noise_injection}).

\subsection{Asynchronous Learning across Reward Dimensions}
\label{sec:async_curves}

\nolinenumbers
\begin{figure}[t]
    \centering
    \includegraphics[width=\columnwidth]{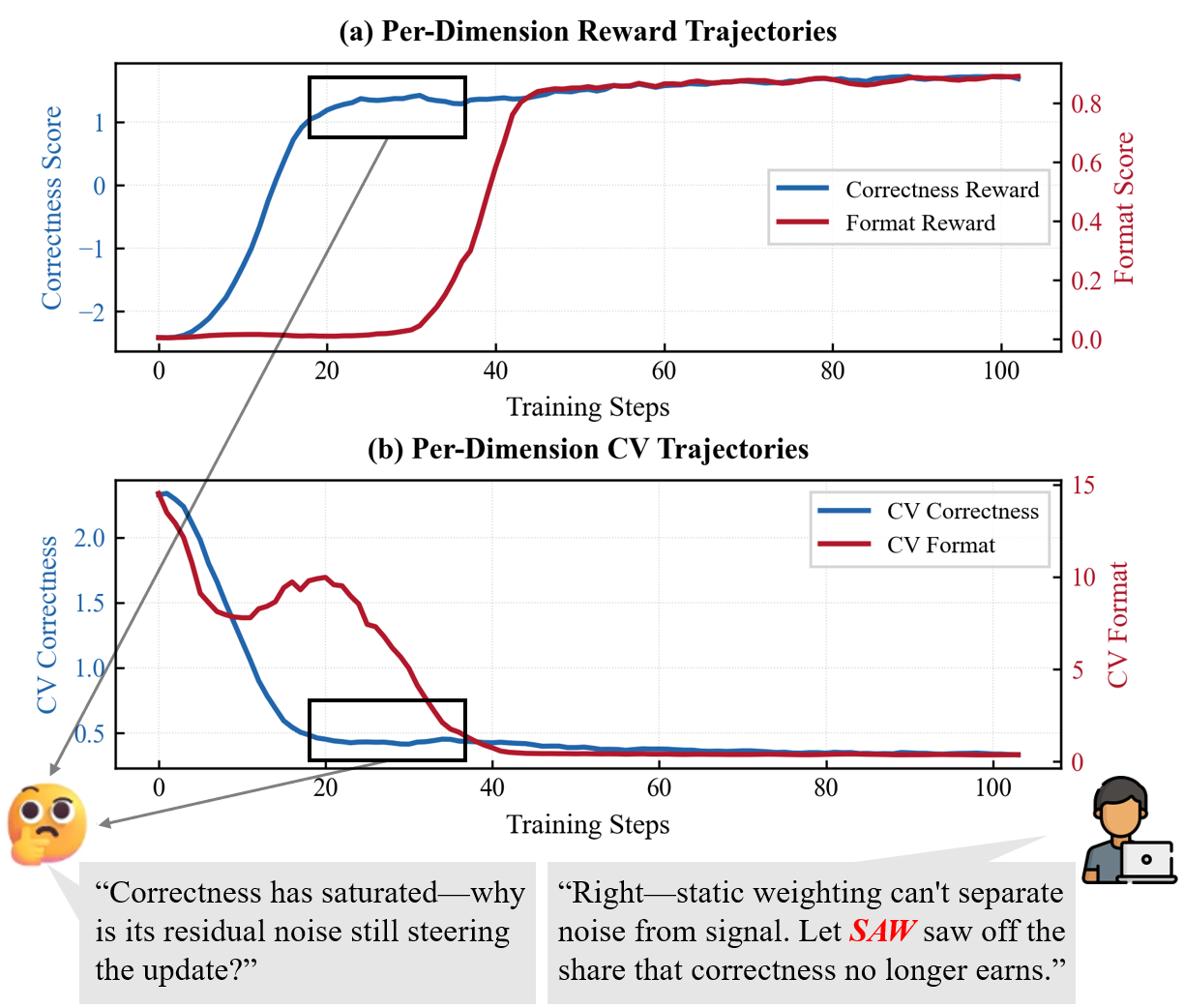}
    \caption{Training dynamics under standard GRPO with equal-weighted aggregation on ToolRL. \textbf{(a)} Per-dimension reward trajectories. The correctness reward saturates rapidly while format continues to evolve. \textbf{(b)} Per-dimension CV trajectories. The CV of the correctness dimension collapses to near zero early in training, while the format CV remains substantially higher throughout most of training.}
    \label{fig:saturation_trap}
\end{figure}
\linenumbers

Figure~\ref{fig:saturation_trap}(a) shows the mean reward trajectories for the two reward dimensions.
The correctness reward rises sharply in the early stages and plateaus around step 25--30, indicating that the model captures the bulk of this dimension's gradient signal within a small fraction of training.
In contrast, the format reward exhibits a slower, more gradual ascent: it remains near zero while correctness is still rising, only entering its rapid-learning phase after correctness has largely saturated, and reaching its own plateau substantially later in training.

This divergence in learning speed directly manifests in the CV trajectories (Figure~\ref{fig:saturation_trap}(b)).
The CV of the correctness dimension drops sharply early in training, signaling that within each batch the correctness dimension now carries limited information and can no longer provide an effective learning signal for the gradient update. Meanwhile, the format CV remains elevated for considerably longer: a high-CV dimension marks a \emph{valuable} learning signal that the gradient update can leverage to reinforce the better samples and suppress the worse ones within each batch of rollouts.

The CV trajectories establish that, by the boxed region (steps $\sim$20--35), correctness has approached convergence ($\mathrm{CV} \to 0$) while format remains the only dimension carrying substantial within-batch variability. In \S\ref{sec:noise_injection}, we analyze the concrete cost this asynchrony imposes on the policy gradient update under static weighting through an example, which isolates the underlying mechanism and shows that the same root cause produces similar failure modes in GRPO and GDPO.

\subsection{How Asynchrony Injects Noise into the Gradient Update}
\label{sec:noise_injection}

Once a reward dimension approaches convergence, its within-batch deviations are stochastic residuals rather than genuine quality differences. Under static-weight aggregation, these residuals enter the policy update through different channels in GRPO and GDPO, but with the same consequence: the gradient direction computed from the advantage is biased. Figure~\ref{fig:noise_injection} illustrates both failure modes on minimal four-rollout examples.

GRPO directly operates on each rollout's absolute deviation $r_{\text{sum}}^{(j)} - \overline{r_{\text{sum}}}$. A near-converged dimension's residual deviation, while small in CV, can match the absolute magnitude of a still-learning dimension's true signal, particularly when the latter is bounded to a small range. The left panel of Figure~\ref{fig:noise_injection} illustrates this concretely: dim 1 is saturated near $1.0$ ($\mathrm{CV} \approx 2\%$) while dim 2 is just beginning to differentiate (with reward $0.04$ on rollout $o_1$), so the absolute magnitudes of dim 1's residual noise ($\pm 0.03$) and dim 2's true signal ($\pm 0.03$ on $o_1$) are comparable. Summing before the group normalization therefore cancels the signal at $o_1$ ($A_{o_1}{=}0$) and elevates the noise-driven $o_2$ to the top rank within the group ($A_{o_2}{=}{+}1.63$).

GDPO instead independently normalizes each individual dimension to unit-variance $z$-scores before final aggregation, so each dimension's contribution depends only on its rank pattern, not its magnitude. A near-converged dimension's negligible fluctuations are thus inflated to $z$-scores of order $\pm 1$, on par with genuine learning signal. The right panel of Figure~\ref{fig:noise_injection} illustrates this: even when dim 2's signal is strong enough that GRPO handles the case correctly, per-dimension $z$-normalization rescales dim 1's small noise ($\pm 0.05$) to $\pm 1.41$, putting it on equal footing with dim 2's true $\pm 1.73$, and the rank between $o_1$ and $o_2$ flips.

\paragraph{A unified view.}
Both failures share the same root cause (static weights cannot distinguish a near-converged dimension's residual noise from a still-learning one's genuine signal), differing only in the noise's entry channel. The same remedy fixes both: weight each dimension's contribution by its CV, applied at GRPO's reward level or GDPO's advantage level. We formalize this next.

\nolinenumbers
\begin{figure*}[t]
    \centering
    \includegraphics[width=\textwidth]{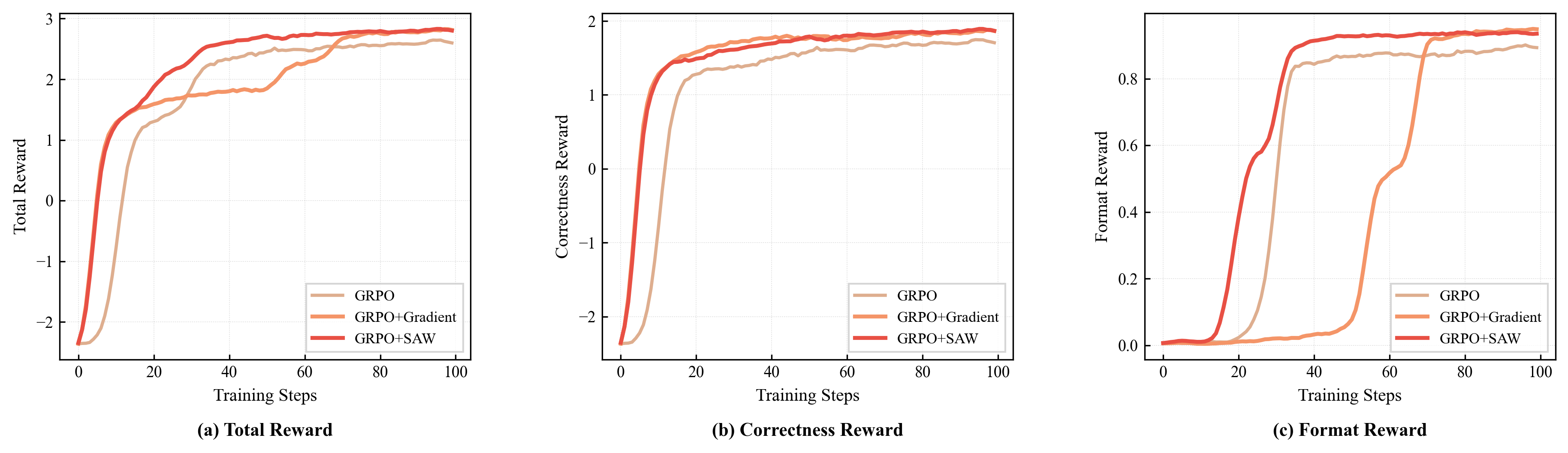}
    \caption{Training trajectories on ToolRL with Qwen2.5-1.5B-Instruct. \method{} accelerates format saturation (the underlearned, high-CV dimension) while leaving correctness convergence essentially unchanged.}
    \label{fig:reward_comparison}
\end{figure*}
\linenumbers

\section{Methodology: Stage-Aware Dynamic Weighting}

\subsection{CV-Based Dynamic Weighting}

As established in \S\ref{sec:prelim}, the coefficient of variation $\cv_k$ (Eq.~\ref{eq:cv}) provides a scale-invariant measure of within-batch reward dispersion, which \S\ref{sec:motivation} shows directly tracks the informativeness of each reward dimension. We compute CV at the batch level rather than per-group, so that statistics average out the noise from individual question difficulty and yield a more stable estimate of the model's global learning state across reward dimensions.

\paragraph{Why CV over other dispersion measures.} Raw variance or standard deviation cannot be directly compared across reward dimensions with different scales: a dimension defined on a wider numerical range will naturally yield a larger standard deviation even when its relative within-batch dispersion is no greater than that of a dimension on a narrower range. CV resolves this by normalizing each dimension's dispersion by its own mean, yielding a dimensionless quantity that reflects \emph{relative} variability. Additionally, CV requires only first and second moments, introducing negligible additional computational overhead per training step.

\subsection{Algorithm Overview}

Given $n$ reward dimensions, we define the base weight for dimension $k$ as its proportional share of total CV within the current batch:
\begin{equation}
    \omega_k' = \frac{\cv_k}{\sum_{l=1}^{n} \cv_l}
    \label{eq:weight}
\end{equation}
For reward-level weighting (GRPO), we use $\omega_k = \omega_k'$ directly, since GRPO's subsequent group-wise normalization (Eq.~\ref{eq:grpo_adv}) absorbs any constant scaling.
For advantage-level weighting (GDPO), we scale by $n$: $\omega_k = n \cdot \omega_k'$, because each per-reward advantage $A_k$ in GDPO is already normalized to zero-mean unit-variance; without the $\times n$ factor, the aggregated advantage would have a systematically smaller magnitude than the standard GDPO formulation with equal weights.

We present the complete \method{} procedure in Algorithm~\ref{alg:sard}. Because the weights are derived directly from current batch statistics, \method{} adapts to each dimension's learning state without any auxiliary state across steps or learnable parameters.

\begin{algorithm}[t]
\caption{\method{}: Stage-Aware Dynamic Weighting for Multi-Reward RL}
\label{alg:sard}
\begin{algorithmic}[1]
\REQUIRE Batch rewards $\{r_k^{(j)}\}_{k=1,\ldots,n}^{j=1,\ldots,N}$; stability constant $\delta$; mode $\in \{\text{reward}, \text{advantage}\}$
\ENSURE Dynamic weights $\{\omega_k\}_{k=1}^{n}$

\STATE \textbf{// Offset non-positive rewards}
\FOR{$k = 1$ \TO $n$}
    \STATE $\tilde{r}_k^{(j)} \leftarrow r_k^{(j)} - r_k^{\min} + \delta, \quad \forall\, j$
\ENDFOR

\STATE \textbf{// Compute batch-level CV for each reward}
\FOR{$k = 1$ \TO $n$}
    \STATE $\mu_k \leftarrow \text{mean}\{\tilde{r}_k^{(1)}, \ldots, \tilde{r}_k^{(N)}\}$
    \STATE $\sigma_k \leftarrow \text{std}\{\tilde{r}_k^{(1)}, \ldots, \tilde{r}_k^{(N)}\}$
    \STATE $\cv_k \leftarrow \sigma_k \,/\, (\mu_k + \delta)$
\ENDFOR

\STATE \textbf{// Compute proportional weights}
\STATE $S \leftarrow \sum_{k=1}^{n} \cv_k$
\IF{$S < \delta$}
    \STATE $\omega_k \leftarrow 1, \quad \forall\, k$ \COMMENT{Fallback to equal weights}
\ELSE
    \FOR{$k = 1$ \TO $n$}
        \STATE $\omega_k \leftarrow \cv_k / S$
        \IF{mode = advantage}
            \STATE $\omega_k \leftarrow \omega_k \times n$
        \ENDIF
    \ENDFOR
\ENDIF

\RETURN $\{\omega_k\}_{k=1}^{n}$
\end{algorithmic}
\end{algorithm}

\subsection{Integration with GRPO and GDPO}
\label{sec:integration}

\method{} is designed as a modular plug-in component with two complementary integration modes:

\paragraph{Mode 1: Reward-level weighting (with GRPO).} \method{} applies dynamic weights directly to the raw rewards before summation:
\begin{equation}
    r_{\text{sum}}^{(j)} = \omega_1\, r_1^{(j)} + \omega_2\, r_2^{(j)} + \cdots + \omega_n\, r_n^{(j)}
    \label{eq:sard_grpo}
\end{equation}
The resulting weighted sum is then processed through GRPO's standard group-wise normalization (Eq.~\ref{eq:grpo_adv}). Since GRPO normalizes the aggregate, the $\times n$ scaling is unnecessary here; only the \emph{relative} weight proportions ultimately matter.

\paragraph{Mode 2: Advantage-level weighting (with GDPO).} Each reward is first independently normalized (Eq.~\ref{eq:gdpo_per_reward}), and then the $\times n$-scaled dynamic weights are applied to the normalized advantages:
\begin{equation}
\begin{split}
    A_{\text{sum}}^{(j)} = \omega_1\, A_1^{(j)} + \omega_2\, A_2^{(j)} \\
    + \cdots + \omega_n\, A_n^{(j)}
\end{split}
    \label{eq:sard_gdpo}
\end{equation}
followed by GDPO's batch-wise normalization (Eq.~\ref{eq:gdpo_batchnorm}). This mode combines GDPO's information-preserving normalization with \method{}'s informativeness-aware weighting, addressing both the reward collapse and the asynchronous reward learning problems simultaneously.

\subsection{Practical Considerations}
\label{sec:practical}

\paragraph{Reward offset.} CV is well-defined only when $\mu_k > 0$. For reward dimensions that may take non-positive values, we apply the offset $\tilde{r}_k^{(j)} = r_k^{(j)} - r_k^{\min} + \delta$ before computing CV, using the theoretical minimum $r_k^{\min}$. The offset rescales each CV but preserves the relative ordering across dimensions, which is sufficient for \method{}'s proportional weighting. For learned reward models without known bounds, empirical batch minima can alternatively be substituted (see Limitations).

\paragraph{Numerical stability.} The $\delta$ term in Eq.~\ref{eq:cv} prevents division by zero when $\mu_k \approx 0$; when all CV values are near zero ($S < \delta$), \method{} falls back to equal weights. \method{} is stateless: weights are computed independently at each step from the current batch statistics, with no cross-step state.

\paragraph{Compatibility with priority weights.} \method{} composes with static priority weights $\alpha_k$ by simple multiplication: $r_{\text{sum}}^{(j)} = \sum_{k} \omega_k\, \alpha_k\, r_k^{(j)}$. Since CV is scale-invariant ($\cv(\alpha_k r_k) = \cv(r_k)$), $\alpha_k$ does not interfere with \method{}'s informativeness assessment. Practitioners control objective priorities through $\alpha_k$, while \method{} independently adjusts allocation by real-time informativeness.

\section{Experiments}

\subsection{Experimental Setup}

We evaluate \method{} on two multi-reward RL tasks with distinct reward structures. Experiments use Qwen2.5-1.5B-Instruct and Qwen2.5-3B-Instruct \citep{qwen25} as base models, trained using the verl framework \citep{verl}. Detailed hyperparameter configurations for both tasks are provided in Appendix~\ref{app:hyperparams}.

\noindent\textbf{ToolRL.} Following \citet{toolrl} and \citet{liu2025gdpo}, we train the model on a tool-calling task with two reward dimensions: a \textbf{format reward} $R_{\text{format}} \in \{0, 1\}$ checking structural compliance, and a \textbf{correctness reward} $R_{\text{correct}} \in [-3, 3]$ evaluating tool calls against ground truth. We adopt the same training set as ToolRL, consisting of 4k samples in total: 2k from ToolACE \citep{toolace}, 1k from Hammer \citep{hammer}, and 1k from xLAM \citep{xlam}. Each instance contains a user question and its corresponding ground-truth tool calls. We evaluate the trained models on the Berkeley Function Call Leaderboard (BFCL-v4) \citep{bfcl}, which spans Non-Live (1{,}150 cases), Live (1{,}381 cases), and Multi-Turn (800 cases) function-calling subsets; a full breakdown of the included sub-tasks is provided in Appendix~\ref{app:bfcl}. The evaluation is run 4 times with the per-instance scores averaged across runs. The training prompt format and reward formulations are given in Appendices~\ref{app:prompt} and \ref{app:reward}.

\noindent\textbf{Reddit TL;DR.} We train Qwen2.5-1.5B-Instruct on a summarization task using the Reddit TL;DR dataset \citep[\texttt{trl-lib/tldr};][]{volske2017tldr,stiennon2020summarize}, randomly sampling 2{,}560 for training and 256 for evaluation. We use two rubric-based reward dimensions: \textbf{quality} (0--10) and \textbf{conciseness} (0--10), scored by Gemini-3-flash-preview-nothinking \citep{gemini} (temperature 0.1); our rubric design follows recent practice of rubric-based reward modeling for RL \citep{huang2025rubricanchors, gunjal2025rubrics, bi2025reward}. The same scoring procedure is applied to both training and evaluation, and the evaluation is run 4 times with the per-instance scores averaged across runs. The training prompt format is provided in Appendix~\ref{app:reddit}, and the full scoring rubric template in Appendix~\ref{app:judge}.

\paragraph{Baselines.} We compare the following methods:
\begin{itemize}
    \item \textbf{Base}: The pre-trained model.
    \item \textbf{GRPO} \citep{deepseekmath}: Multi-reward GRPO with equal-weight reward summation.
    \item \textbf{GRPO+Gradient} \citep{lu2025dynamic}: GRPO with gradient-based dynamic reward weighting. At each training step, per-objective gradients are computed and their pairwise dot products form an interaction matrix; weights are updated via exponentiated gradient ascent to favor dimensions whose gradients align with the overall optimization direction.
    \item \textbf{GRPO+\method{}}: GRPO with CV-based dynamic reward weighting (Mode 1).
    \item \textbf{GDPO} \citep{liu2025gdpo}: Decoupled normalization with equal-weight advantage summation across reward dimensions.
    \item \textbf{GDPO+Gradient}: GDPO with gradient-based dynamic advantage weighting.
    \item \textbf{GDPO+\method{}}: GDPO with CV-based dynamic advantage weighting (Mode 2).
\end{itemize}

\begin{table}[t]
    \centering
    \resizebox{\columnwidth}{!}{%
    \begin{tabular}{l ccc}
        \toprule
        \textbf{Method} & \textbf{Total $\uparrow$} & \textbf{Quality $\uparrow$} & \textbf{Conc. $\uparrow$} \\
                         & (/20) & (/10) & (/10) \\
        \midrule
        \textit{Qwen2.5-1.5B-Instruct} & \textit{14.69} & \textit{7.34} & \textit{7.35} \\
        \midrule
        GRPO              & 18.20          & 8.71          & 9.49 \\
        GRPO+Gradient     & 18.76\small{$_{+.56}$} & 9.11\small{$_{+.40}$} & 9.64\small{$_{+.15}$} \\
        GRPO+\method{}    & \textbf{18.80}\small{$_{+.60}$} & \textbf{9.15}\small{$_{+.44}$} & \textbf{9.65}\small{$_{+.16}$} \\
        \midrule
        GDPO              & 18.80          & 9.10          & 9.70 \\
        GDPO+Gradient     & 18.94\small{$_{+.14}$} & 9.17\small{$_{+.07}$} & \textbf{9.77}\small{$_{+.07}$} \\
        GDPO+\method{}    & \textbf{19.07}\small{$_{+.27}$} & \textbf{9.35}\small{$_{+.25}$} & 9.72\small{$_{+.02}$} \\
        \bottomrule
    \end{tabular}%
    }
    \caption{Reddit TL;DR results (2 rewards). Quality and Conciseness (Conc.) are scored 0--10 by a rubric-based LLM judge; Total is their sum. Subscripts denote improvements over the corresponding baseline.}
    \label{tab:reddit}
\end{table}

\subsection{Evaluation}

\noindent\textbf{ToolRL.} Table~\ref{tab:toolrl} reveals two patterns that hold across both model scales. CV reweighting alone closes most of the gap between GRPO and GDPO, directly supporting the unified noise-injection view of \S\ref{sec:noise_injection}: since decoupled normalization and CV-based reweighting address the same underlying problem from different angles, a fix at either point recovers most of the loss. \method{} also matches or exceeds the gradient-based weighting of \citet{lu2025dynamic} on every aggregate metric at a small fraction of the compute (\S\ref{sec:efficiency}), showing that for multi-objective RLHF, CV-based dynamic weighting is both more effective and substantially lighter than gradient-based dynamic weighting. Beyond aggregate metrics, the training trajectories in Figure~\ref{fig:reward_comparison} make the mechanism concrete: under GRPO+\method{}, the format reward rises sharply once the correctness reward saturates around step 25--30 and reaches its plateau noticeably earlier than under GRPO or GRPO+Gradient, neither of which manages to accelerate format learning in the post-saturation phase. This directly visualizes the mechanism formalized in \S\ref{sec:noise_injection}: once one dimension nears convergence, \method{} reallocates the optimization budget toward the still-learning dimension instead of letting the saturated dimension's residual noise dominate the gradient direction. Composition with GDPO yields the strongest configuration at both 1.5B and 3B, confirming that the two mechanisms are complementary rather than substitutable.

\noindent\textbf{Reddit TL;DR.} Table~\ref{tab:reddit} further supports this picture: both GRPO+\method{} and GDPO+\method{} achieve the top total score within their respective groups, and introducing \method{} improves the scores on \emph{both} the quality and the conciseness dimensions over the corresponding baselines. Unlike the ToolRL setting where one dimension (correctness) saturates within the first few dozen steps and \method{}'s gain is concentrated on accelerating the still-learning format dimension, both Reddit reward dimensions remain in the active-learning regime throughout training; the improvement from \method{} therefore reflects \emph{continuous} reweighting between two evolving signals rather than a transient catch-up effect, raising the attainable performance ceiling on both reward dimensions.

\subsection{Efficiency Analysis}
\label{sec:efficiency}

\method{} and the gradient-based method of \citet{lu2025dynamic} both weight at batch-level per-dimension granularity, but at vastly different cost: the latter requires $n$ extra forward-backward passes per step to form a gradient interaction matrix, $O(nP)$ in model parameters, whereas \method{} needs only batch-level mean and standard deviation, $O(nN)$ in batch size.

\begin{table}[h]
    \centering
    \small
    \resizebox{\columnwidth}{!}{%
    \begin{tabular}{lcccc}
        \toprule
        \textbf{Method} & \textbf{1.5B (s)} & \textbf{+\%} & \textbf{3B (s)} & \textbf{+\%} \\
        \midrule
        GRPO              & 69.49   & --        & 119.26  & --       \\
        GRPO+Gradient     & 179.75  & +158.7\%  & 344.36  & +188.7\% \\
        GRPO+\method{}    & 70.11   & +0.9\%    & 122.43  & +2.7\%   \\
        \midrule
        GDPO              & 71.90   & --        & 123.71  & --       \\
        GDPO+Gradient     & 186.30  & +159.1\%  & 356.67  & +188.3\% \\
        GDPO+\method{}    & 73.10   & +1.7\%    & 124.82  & +0.9\%   \\
        \bottomrule
    \end{tabular}%
    }
    \caption{Average wall-clock time per training step (seconds) on ToolRL across two model scales, with $n{=}2$ reward dimensions. The \textbf{+\%} columns show overhead relative to the corresponding baseline (GRPO or GDPO).}
    \label{tab:efficiency}
\end{table}

Empirically (Table~\ref{tab:efficiency}), the gradient-based method incurs 159--189\% wall-clock overhead while \method{} adds under 3\% across all the settings, with the absolute gap widening at scale: on Qwen2.5-3B, the gradient-based method requires over 344s per step against 119s for GRPO, whereas \method{} adds at most 3.2s. \method{} therefore achieves competitive or even superior reward weighting at a fraction of the computational cost, making it practical for large-scale RLHF training.

\section{Conclusion}

In multi-objective reinforcement learning for large language models, static weighted summation overlooks a key dynamic challenge: reward learning is asynchronous across objectives, and noise from near-converged dimensions contaminates the gradient updates that should be driven by the still-learning ones. We propose \methodfull{} (\method{}), a dynamic weighting mechanism that addresses this asynchrony by reweighting each dimension's reward or advantage contribution according to its coefficient of variation. Relying only on batch-level statistics, \method{} consistently improves multi-objective optimization across two RL frameworks (GRPO and GDPO) and two task families with markedly different reward dynamics. This suggests that CV-based informativeness is a broadly applicable signal rather than a task-specific heuristic. We hope this work inspires further research on informativeness-driven optimization for multi-objective RLHF and on principled weighting mechanisms for aligning LLMs with diverse and evolving human preferences.

\clearpage
\section*{Limitations}

Our method has several limitations. First, the coefficient of variation is not shift-invariant: applying a constant offset to ensure positive means (as required for well-defined CV) compresses the dynamic range of CV values, particularly for reward dimensions with large negative lower bounds. In our experiments, where theoretical bounds are known and the offset is consistent, this does not impair the relative ranking of CV across dimensions; however, for learned reward models without known bounds, alternative strategies such as batch-level empirical minima or min-max normalization prior to CV computation may be more appropriate and warrant further investigation. Second, our experiments focus on settings with two reward dimensions; the behavior of \method{} with a larger number of objectives (e.g., $n \geq 3$) remains to be explored. Third, the batch-level CV computation provides a global estimate of learning state that does not capture per-question variation; exploring finer-grained (e.g., group-level) weighting with appropriate smoothing techniques could improve adaptiveness. Finally, while \method{} is algorithm-agnostic in principle, we have only empirically validated it with GRPO and GDPO; its effectiveness with other RL algorithms (e.g., PPO, REINFORCE variants) requires further study.

\bibliography{custom}

\clearpage
\appendix
\section*{Appendix}

\section{Related Work}
\label{app:related}

\paragraph{GRPO Variants.}
Group Relative Policy Optimization \citep{deepseekmath} has become the de facto algorithm for LLM reinforcement learning, spurring a family of variants that address its limitations from different angles.
GRPO and its variants have since been applied across a broad range of tasks, from mathematical reasoning \citep{deepseekmath, deepseekr1} to tool-use and agentic reasoning \citep{toolrl, tong2026autagent}.
DAPO \citep{dapo} introduces dynamic sampling and decoupled clipping to stabilize training at scale.
Dr.GRPO \citep{drgrpo} provides a critical analysis of R1-Zero-like training and proposes corrections to GRPO's variance reduction properties.
To improve stability, Group Sequence Policy Optimization (GSPO) \citep{gspo} defines the importance ratio based on sequence likelihood rather than at the token level, performing sequence-level clipping, rewarding, and optimization.
To promote efficient reasoning, Group Filtered Policy Optimization (GFPO) \citep{gfpo} addresses length explosion by sampling larger groups per problem during training and filtering responses based on their length and reward-per-token ratio.
These works focus on improving the core optimization dynamics of single-objective or aggregate-reward RL; our work is complementary, targeting how multiple reward signals are \emph{weighted} before entering the optimization loop, and is compatible with any GRPO-family algorithm.

\paragraph{Multi-Objective Reinforcement Learning.}
Multi-objective formulations have been broadly adopted across LLM RL pipelines. At the task level, multi-reward designs have been applied to tool calling \citep{toolrl}, summarization \citep{chen2026reward}, and safety-oriented alignment \citep{mu2024rulebased, chen2026reward}; recent frontier systems also build on this paradigm, with DeepSeek-V3.2 \citep{deepseekv32} combining rule-based outcome rewards, length penalties, and language-consistency rewards to scale reasoning and agentic capabilities.
Another important recent application of multi-reward RL targets \emph{efficient reasoning}, where length-based rewards are paired with outcome rewards to compress chain-of-thought without sacrificing accuracy: O1-Pruner \citep{luo2025o1pruner} and \citet{arora2025efficient} apply normalized length penalties for proportional compression; ShorterBetter \citep{yi2025shorterbetter} promotes conciseness by penalizing deviations from the shortest correct response within a sampled group; L1 \citep{aggarwal2025l1} introduces Length-Controlled Policy Optimization to honor a target length budget; \citet{liu2025adaptive} dynamically adjust the accuracy--length trade-off based on running model performance; and DLER \citep{liu2025dler} revisits the simplest truncation-based length penalty and shows that, with a carefully designed RL recipe (batch-wise reward normalization, higher clipping, and dynamic sampling), it suffices to achieve state-of-the-art accuracy--length trade-offs.

Methodologically, prior work tackles multi-objective trade-offs along three axes.
\emph{Normalization-based} approaches isolate per-dimension signal: GDPO \citep{liu2025gdpo} resolves the reward collapse problem through per-reward decoupled normalization (\S\ref{sec:prelim}), and MO-GRPO \citep{mogrpo} independently arrives at a similar per-reward variance normalization strategy.
\emph{Reward-design} approaches reshape how the objectives combine: Safe RLHF \citep{dai2023safe} balances helpfulness and safety via constrained optimization with separate reward and cost models; ALARM \citep{lai2024alarm} introduces a hierarchical reward structure that filters aspect rewards by their consistency with a holistic preference signal; and Fine-Grained RLHF \citep{wu2023finegrained} decomposes feedback into multiple aspect-specific reward models that fire at sentence- and span-level granularity.
\emph{Pareto-oriented} approaches recast multi-objective RL as direct Pareto-front navigation: PAMA \citep{he2025pareto} converts multi-objective aggregation into a closed-form convex optimization within the PPO framework to converge to a Pareto-stationary point, while GAPO \citep{gapo} adaptively balances policy-gradient directions via multi-gradient descent.
In contrast to all of the above, which operate on \emph{what} signal each dimension contributes, \method{} addresses \emph{when}: it allocates optimization budget across dimensions based on their real-time, batch-level informativeness, and composes orthogonally with these existing mechanisms.
\paragraph{Dynamic Reward Weighting.}
The most directly related work is \citet{lu2025dynamic}, who propose two dynamic weighting methods for multi-objective RL.
Their \emph{hypervolume-guided} method computes a step-level scalar meta-reward based on Pareto front contribution that uniformly scales all reward dimensions; it does not produce per-dimension weights, is not directly comparable to \method{}, and we therefore do not include it as a baseline.
Their \emph{gradient-based} method computes per-objective gradients at each step, forms a gradient interaction matrix via pairwise dot products, and updates per-dimension weights via exponentiated gradient ascent; its granularity matches \method{} (batch-level, per-dimension), but each step requires $n$ additional forward-backward passes, multiplying training cost by $(1+n)$, and depends on custom distributed gradient-collection infrastructure.
By contrast, \method{} achieves comparable per-dimension weighting using only batch-level mean and standard deviation, with negligible overhead and no modification to the training infrastructure.

\nolinenumbers
\begin{figure}[t]
    \centering
    \includegraphics[width=\columnwidth]{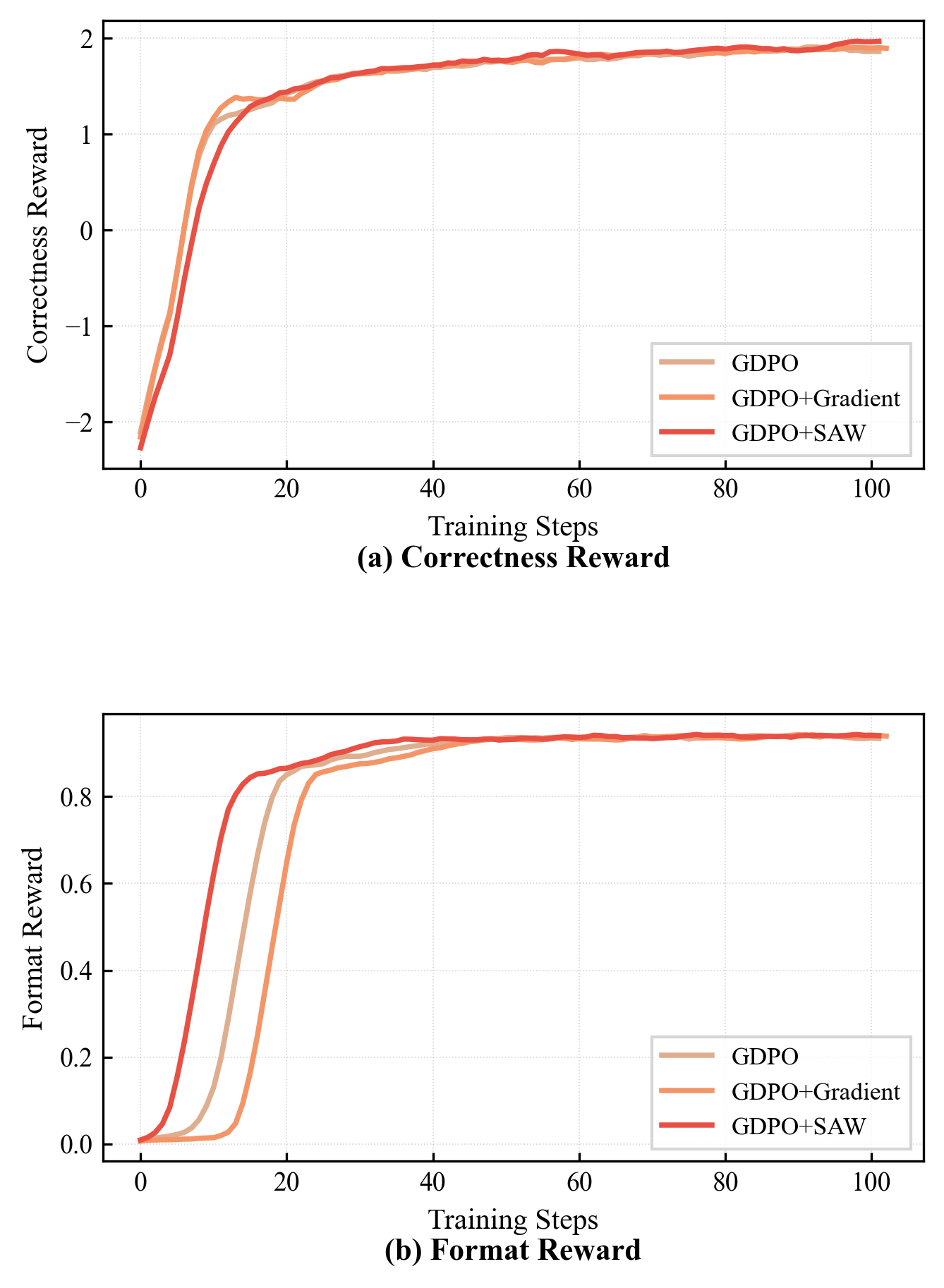}
    \caption{Per-dimension reward trajectories for GDPO-based methods during training on ToolRL with Qwen2.5-1.5B-Instruct. \textbf{(a)} Correctness reward: all three variants converge to similar terminal values. \textbf{(b)} Format reward: GDPO+\method{} reaches the saturation plateau earlier than GDPO and GDPO+Gradient, consistent with \method{}'s tendency to up-weight the high-CV dimension during early training.}
    \label{fig:gdpo_toolrl_curves}
\end{figure}
\linenumbers

\nolinenumbers
\begin{figure}[t]
    \centering
    \includegraphics[width=\columnwidth]{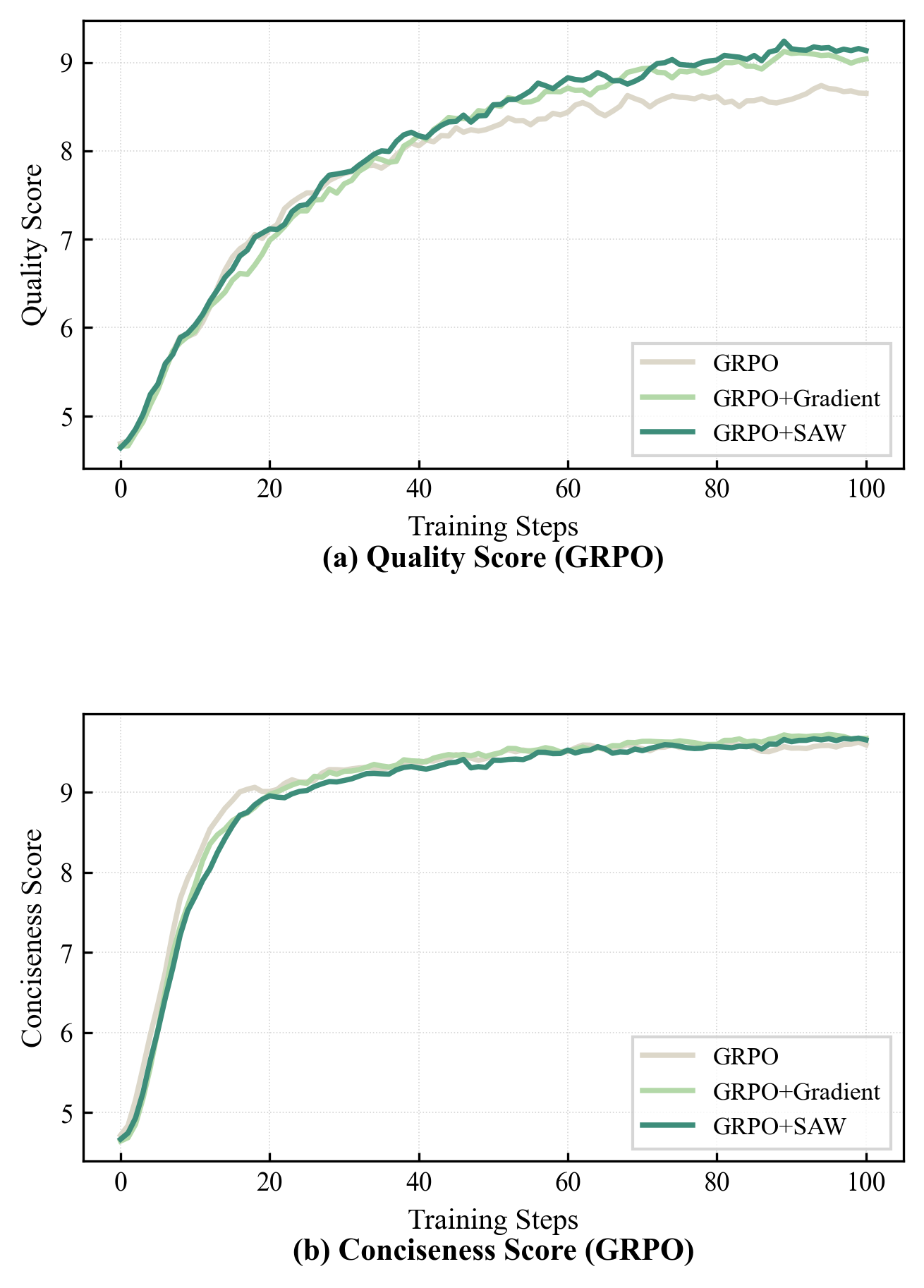}
    \caption{Per-dimension reward trajectories for GRPO-based methods during training on Reddit TL;DR with Qwen2.5-1.5B-Instruct.}
    \label{fig:reddit_reward_grpo}
\end{figure}
\linenumbers

\nolinenumbers
\begin{figure}[t]
    \centering
    \includegraphics[width=\columnwidth]{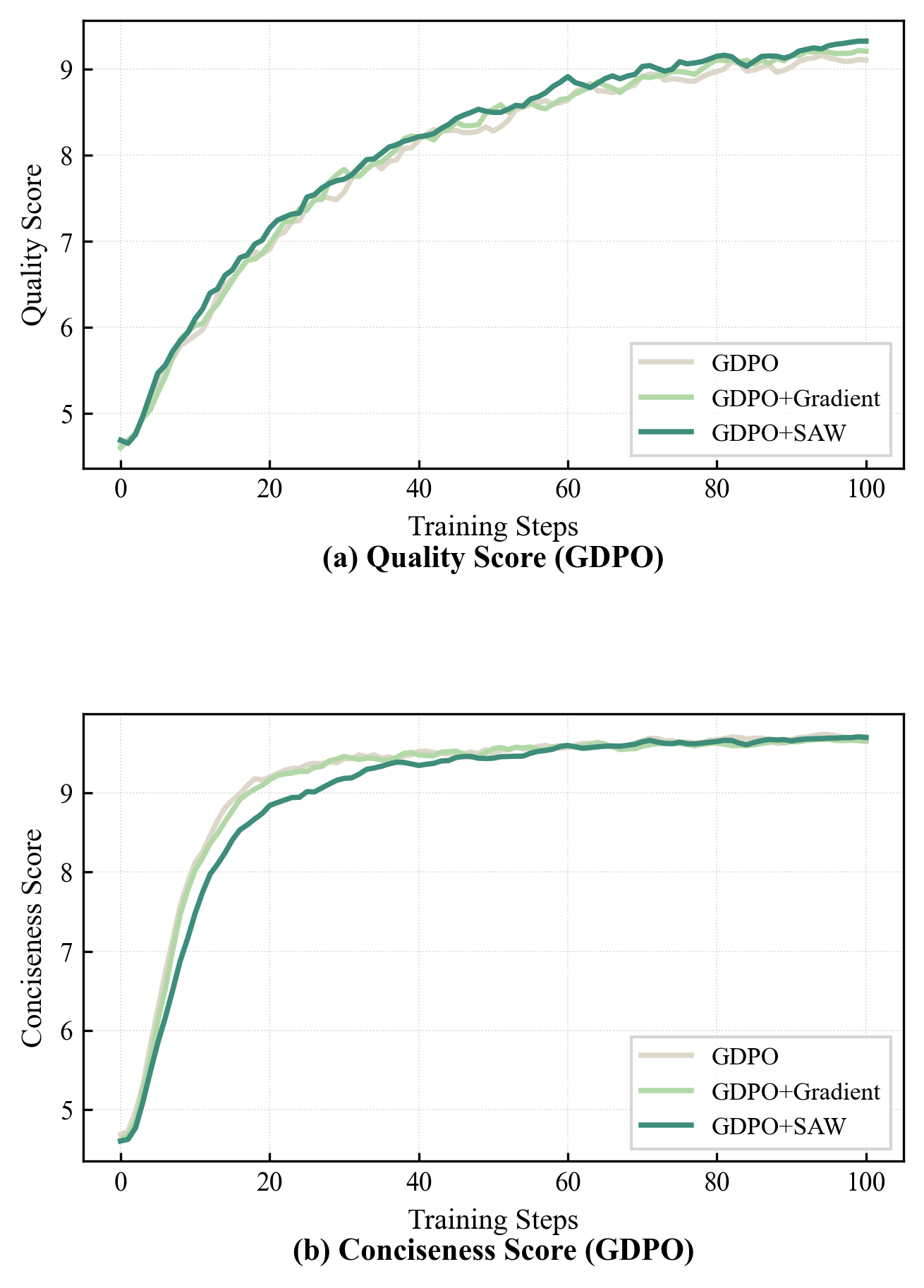}
    \caption{Per-dimension reward trajectories for GDPO-based methods during training on Reddit TL;DR with Qwen2.5-1.5B-Instruct.}
    \label{fig:reddit_reward_gdpo}
\end{figure}
\linenumbers

\section{ToolRL Training Prompt Format}
\label{app:prompt}

We follow the same setup as GDPO and ToolRL; the training prompt is given below.

\begin{tcolorbox}[breakable, colback=blue!5!white, colframe=blue!75!black, title=System Prompt for ToolRL Training]
You are a helpful dialogue assistant capable of leveraging tool calls to solve user tasks and provide structured chat responses.

\textbf{Available Tools}\\
In your response, you can use the following tools:\\
\{\{ Tool List \}\}

\textbf{Steps for Each Turn}
\begin{enumerate}[leftmargin=*]
    \item \textbf{Think}: Recall relevant context and analyze the current user goal.
    \item \textbf{Decide on Tool Usage}: If a tool is needed, specify the tool and its parameters.
    \item \textbf{Respond Appropriately}: If a response is needed, generate one while maintaining consistency across user queries.
\end{enumerate}

\textbf{Output Format}\\
{\color{olive}\texttt{<think>}} Your thoughts and reasoning {\color{olive}\texttt{</think>}}\\
{\color{purple}\texttt{<tool\_call>}} \{"name": "Tool name", "parameters": \{"Parameter name": "Parameter content", "...~...": "...~..."\}\}\\
\{"name": "...~...", "parameters": \{"...~...": "...~...", "...~...": "...~..."\}\}\\
...\\
{\color{purple}\texttt{</tool\_call>}}\\
{\color{teal}\texttt{<response>}}AI's final response {\color{teal}\texttt{</response>}}

\textbf{Important Notes}
\begin{enumerate}[leftmargin=*]
    \item You must always include the {\color{olive}\texttt{<think>}} field to outline your reasoning. Provide at least one of {\color{purple}\texttt{<tool\_call>}} or {\color{teal}\texttt{<response>}}. Decide whether to use {\color{purple}\texttt{<tool\_call>}} (possibly multiple times), {\color{teal}\texttt{<response>}}, or both.
    \item You can invoke multiple tool calls simultaneously in the {\color{purple}\texttt{<tool\_call>}} fields. Each tool call should be a JSON object with a ``name'' field and a ``parameters'' field containing a dictionary of parameters. If no parameters are needed, leave the ``parameters'' field an empty dictionary.
    \item Refer to the previous dialogue records in the history, including the user's queries, previous {\color{purple}\texttt{<tool\_call>}}, {\color{teal}\texttt{<response>}}, and any tool feedback noted as {\color{orange}\texttt{<obs>}} (if exists).
\end{enumerate}
\end{tcolorbox}

\begin{tcolorbox}[breakable, colback=yellow!5!white, colframe=blue!75!black, title=User Prompt for ToolRL Training]
\textbf{Dialogue History}

\texttt{<user>} \{\{ Initial User Input \}\} \texttt{</user>}

{\color{olive}\texttt{<think>}}~~Round 1 Model Thought~~{\color{olive}\texttt{</think>}}\\
\{\{ Round 1 model output {\color{purple}\texttt{<tool\_call>}} or {\color{teal}\texttt{<response>}} \}\}\\
{\color{orange}\texttt{<obs>}} Round 1 Observation {\color{orange}\texttt{</obs>}}

...~...

\texttt{<user>} \{\{ User Input \}\} \texttt{</user>}

...~...
\end{tcolorbox}

\section{Tool Calling Reward Functions}
\label{app:reward}

\paragraph{Format Reward.} The format reward $\mathcal{R}_{\text{format}} \in \{0, 1\}$ checks whether the model output satisfies the required structure and contains all necessary fields in the correct order:
\begin{equation}
\mathcal{R}_{\text{format}} = \begin{cases} 1, & \text{correct structure,} \\ 0, & \text{otherwise.} \end{cases}
\end{equation}

\paragraph{Correctness Reward.} The correctness reward $\mathcal{R}_{\text{correct}} \in [-3,\, 3]$ evaluates the predicted tool calls $P = \{P_1, \ldots, P_m\}$ against the ground-truth calls $G = \{G_1, \ldots, G_n\}$. It consists of three components:

\begin{itemize}[leftmargin=*]
\item \textbf{Tool Name Matching:}
\begin{equation}
r_{\text{name}} = \frac{|N_G \cap N_P|}{|N_G \cup N_P|} \in [0, 1],
\end{equation}
where $N_G$ and $N_P$ are the sets of tool names from ground-truth and predicted calls, respectively.

\item \textbf{Parameter Name Matching:}
\begin{equation}
r_{\text{param}} = \!\sum_{G_j \in G} \frac{|\mathrm{keys}(G_j) \cap \mathrm{keys}(P_j)|}{|\mathrm{keys}(G_j) \cup \mathrm{keys}(P_j)|}
\end{equation}
$r_{\text{param}} \in [0, |G|]$, where $\mathrm{keys}(\cdot)$ denotes the parameter names of a call.

\item \textbf{Parameter Content Matching:}
\begin{equation}
r_{\text{value}} = \!\!\sum_{G_j \in G}\; \sum_{k \in \mathrm{keys}(G_j)} \!\!\mathbf{1}[P_G[k] = P_P[k]]
\end{equation}
$r_{\text{value}} \in [0,\, \sum_{G_j \in G} |\mathrm{keys}(G_j)|]$, where $P_G[k]$ and $P_P[k]$ are the parameter values for the ground-truth and predicted calls.

\item \textbf{Total Match Score:}
\begin{equation}
r_{\text{match}} = r_{\text{name}} + r_{\text{param}} + r_{\text{value}}
\end{equation}
where $r_{\text{match}} \in [0, S_{\max}]$ and
\begin{equation}
S_{\max} = 1 + |G| + \!\sum_{G_j \in G} |\mathrm{keys}(G_j)|.
\end{equation}
\end{itemize}

The final correctness reward is computed by finding the optimal matching between $P$ and $G$ to maximize the total match score:
\begin{equation}
\mathcal{R}_{\text{correct}} = 6 \cdot \frac{R_{\max}}{S_{\max}} - 3 \in [-3,\, 3],
\end{equation}
where $R_{\max}$ denotes the total match score from the optimal matching.

\section{ToolRL Hyperparameters Setting}
\label{app:hyperparams}

All hyperparameter settings are kept identical to those used in ToolRL \citep{toolrl} and GDPO \citep{liu2025gdpo}. The complete training configuration is listed in Table~\ref{tab:hyperparams}.

\begin{table}[h]
\centering
\small
\resizebox{\columnwidth}{!}{%
\begin{tabular}{ll}
\toprule
\textbf{Parameter} & \textbf{Value} \\
\midrule
Total Epochs & 15 \\
Train Batch Size & 512 \\
Mini Batch Size & 128 \\
Max Prompt Length & 2048 \\
Max Response Length & 1024 \\
Learning Rate & 1e-6 \\
Group Size ($G$) & 4 \\
KL Coefficient & 0.001 \\
KL Loss & Disabled \\
Gradient Checkpointing & Enabled \\
Rollout Engine & vLLM \\
Rollout TP Size & 1 \\
GPU Memory Utilization & 0.6 \\
Number of GPUs & $4\times$ NVIDIA A100 80GB \\
\bottomrule
\end{tabular}%
}
\caption{ToolRL training configuration using the verl framework. All hyperparameter settings follow ToolRL \citep{toolrl}.}
\label{tab:hyperparams}
\end{table}

\section{BFCL-v4 Subsets}
\label{app:bfcl}

We evaluate on the Berkeley Function Call Leaderboard v4 (BFCL-v4), which is organized into three top-level categories. Each category contains multiple sub-tasks, listed in Table~\ref{tab:bfcl}; the reported accuracies in Table~\ref{tab:toolrl} are the macro averages over these sub-tasks within each category.

\begin{table}[h]
\centering
\small
\begin{tabular}{lll}
\toprule
\textbf{Category} & \textbf{\#Cases} & \textbf{Sub-tasks} \\
\midrule
Non-Live & 1{,}150 & \texttt{simple\_python}, \\
        &         & \texttt{simple\_java}, \\
        &         & \texttt{simple\_javascript}, \\
        &         & \texttt{multiple}, \\
        &         & \texttt{parallel}, \\
        &         & \texttt{parallel\_multiple} \\
\midrule
Live    & 1{,}381 & \texttt{live\_simple}, \\
        &         & \texttt{live\_multiple}, \\
        &         & \texttt{live\_parallel}, \\
        &         & \texttt{live\_parallel\_multiple} \\
\midrule
Multi-Turn & 800  & \texttt{multi\_turn\_base}, \\
        &         & \texttt{multi\_turn\_miss\_func}, \\
        &         & \texttt{multi\_turn\_miss\_param}, \\
        &         & \texttt{multi\_turn\_long\_context} \\
\bottomrule
\end{tabular}
\caption{Composition of BFCL-v4 used in our evaluation.}
\label{tab:bfcl}
\end{table}

\section{ToolRL Training Curves under GDPO}
\label{app:gdpo-curves}

Figure~\ref{fig:gdpo_toolrl_curves} reports the per-dimension reward trajectories of GDPO, GDPO+Gradient, and GDPO+\method{} on ToolRL with Qwen2.5-1.5B-Instruct, complementing the GRPO-based curves in Figure~\ref{fig:reward_comparison}. All three variants converge to comparable terminal levels on both reward dimensions; \method{} primarily affects the early-to-mid phase, accelerating format reward saturation while maintaining correctness improvement. This is consistent with the BFCL-v4 results in Table~\ref{tab:toolrl}, where the gains of \method{} over GDPO at convergence are modest because both reward dimensions have largely saturated in this setting; the more substantial gains appear in the non-saturating Reddit TL;DR setting (Table~\ref{tab:reddit}).

\section{Reddit TL;DR Training Prompt Format}
\label{app:reddit}

The system prompt used during training is shown below, followed by one representative example from the Reddit TL;DR dataset.

\begin{tcolorbox}[breakable, colback=blue!5!white, colframe=blue!75!black, title=System Prompt for Reddit TL;DR Training]
You are a professional content editor. Please read the following post and summarize its main message in concise and accurate language.
\end{tcolorbox}

\begin{tcolorbox}[breakable, colback=yellow!5!white, colframe=blue!75!black, title=User Prompt for Reddit TL;DR Training]
SUBREDDIT: r/AskReddit

TITLE: Java question getting desperate!

POST: I am currently enrolled in a ``Java 2'' Class in college. And my current project is to write a program that can solve mazes using two different methods 1. Depth First Search using a stack and 2. Breadth First Search using a queue. I have depth first search done, and have yet to begin working on breadth first search but i have a few ideas as to how to implement it. My problem is, with the client/driver program that runs all of the classes (contains my main method). I have to scan in text files that contain mazes. I have tested my code by scanning the mazes and everything is working fine. The trouble comes with the instructor says that he wants to be able to put a directory into the command line (args) and have our program parse the directory and solve each maze in the directory. I did directory.list() and place all of the file names in the directory into an array. But I do not know how to make it so my program can scan those files. The program gets upset because the files are not in the same directory as the code. If anyone can help with this I would forever be in your debt.

TL;DR:
\end{tcolorbox}

\section{Reddit TL;DR LLM-as-Judge Rubric}
\label{app:judge}

We use Gemini-3-flash-preview-nothinking (temperature 0.1) as an LLM judge to score generated summaries. Since the Reddit TL;DR dataset includes human-written reference summaries, we provide both the original post and the human reference to the judge, enabling more calibrated scoring. The full scoring prompt is shown below.

\begin{tcolorbox}[breakable, colback=blue!5!white, colframe=blue!75!black, title={\large LLM-as-Judge Scoring Prompt}]
\normalsize
\textbf{\# Role}\\
You are a highly rigorous linguistic auditor. Your task is to perform a comparative evaluation of a ``Model-Generated Summary'' against a source text and a reference summary.

\vspace{0.5em}
\textbf{\# Evaluation Dimensions}

\vspace{0.3em}
\textbf{\#\# 1. Quality (Focus: Accuracy \& Coverage)}\\
Score this dimension based on factual correctness and essential information retention.
\begin{itemize}[leftmargin=*, nosep]
\item 9--10 (Perfect): No hallucinations. Captures all core entities and the main conflict/conclusion mentioned in the human reference.
\item 7--8 (Minor Issue): Factual alignment is 100\%, but misses one secondary supporting detail from the human reference.
\item 5--6 (Major Omission): Misinterprets a secondary detail OR misses the primary ``hook''/conclusion of the post.
\item 3--4 (Severe Error): Contains at least one significant factual hallucination OR fails to address the main subject.
\item 1--2 (Failure): Completely irrelevant, toxic, or incoherent.
\end{itemize}

\vspace{0.3em}
\textbf{\#\# 2. Conciseness (Focus: Word Efficiency \& Redundancy)}\\
Score this dimension based on the ``Information-to-Word Ratio.''
\begin{itemize}[leftmargin=*, nosep]
\item 9--10 (Optimal): Zero filler phrases. Every word contributes to the meaning. No repetition.
\item 7--8 (Slightly Wordy): Includes 1--2 unnecessary words or ``meta-talk'' (e.g., ``The user says...'', ``In this story...'').
\item 5--6 (Redundant): Contains repetitive information or excessive adjectives that do not add meaning.
\item 3--4 (Inefficient): Is longer than the human reference without adding extra useful info, OR 30\% of the text is filler.
\item 1--2 (Bloated): Verbose, contains circular reasoning, or is nearly the same length as the original post.
\end{itemize}

\vspace{0.5em}
\textbf{\# Execution Logic (Chain of Thought)}
\begin{enumerate}[leftmargin=*, nosep]
\item Identify the ``Core Message'' from the Human Summary.
\item Check if the Model Summary contains this Core Message (Quality).
\item Count ``Filler Words'': phrases that can be removed without changing the meaning (Conciseness).
\item Penalize hallucinations heavily.
\end{enumerate}

\vspace{0.5em}
\textbf{\# Task Input}\\[0.3em]
{[Original Reddit Post]:}\\
\texttt{\$reddit\_text}\\[0.3em]
{[Human-Written Summary (Reference)]:}\\
\texttt{\$human\_summary}\\[0.3em]
{[Model-Generated Summary (To be evaluated)]:}\\
\texttt{\$model\_summary}

\vspace{0.5em}
\textbf{\# Constraints}
\begin{enumerate}[leftmargin=*, nosep]
\item Output ONLY a valid JSON object.
\item Be clinical and objective.
\end{enumerate}

\vspace{0.5em}
\textbf{\# Output Format}
\begin{verbatim}
{
  "quality": {
    "score": [Integer]
  },
  "conciseness": {
    "score": [Integer]
  }
}
\end{verbatim}
\end{tcolorbox}

\section{Reddit TL;DR Hyperparameters Setting}
\label{app:reddit-hyperparams}

We list below the hyperparameter setting used in our Reddit TL;DR experiments. The complete training configuration is given in Table~\ref{tab:reddit-hyperparams}.

\begin{center}
\small
\refstepcounter{table}\label{tab:reddit-hyperparams}
\begin{tabular}{ll}
\toprule
\textbf{Parameter} & \textbf{Value} \\
\midrule
Total Epochs & 11 \\
Train Batch Size & 256 \\
Mini Batch Size & 64 \\
Max Prompt Length & 2048 \\
Max Response Length & 1024 \\
Learning Rate & 1e-6 \\
Group Size ($G$) & 4 \\
KL Coefficient & 0.001 \\
KL Loss & Disabled \\
Gradient Checkpointing & Enabled \\
Rollout Engine & vLLM \\
Rollout TP Size & 1 \\
GPU Memory Utilization & 0.6 \\
Number of GPUs & $4\times$ NVIDIA A100 80GB \\
\bottomrule
\end{tabular}

\vspace{0.4em}
\small Table~\thetable: Reddit TL;DR training configuration using the verl framework.
\end{center}

\section{Reddit TL;DR Training Curves}
\label{app:curves}

The complete training curves for the Reddit TL;DR experiments are shown in Figures~\ref{fig:reddit_reward_grpo} and \ref{fig:reddit_reward_gdpo}.

\end{document}